Paper:

# Frailty Care Robot for Elderly and its Application for Physical and Psychological Support

**Yoichi Yamazaki**[*1,†], **Masayuki Ishii**[*2], **Takahiro Ito**[*3], **and Takuya Hashimoto**[*4]

[*1]Department of Home Electronics, Faculty of Creative Engineering, Kanagawa Institute of Technology
1030 Shomo-Ogino, Atsugi, Kanagawa 243-0292, Japan
E-mail: yamazaki@he.kanagawa-it.ac.jp
[*2]OKUBO GEAR Co., Ltd.
3030 Kamiechi, Atsugi, Kanagawa 243-0801, Japan
[*3]CMK Products Corporation
1-1-11 Tanashioda, Chuo, Sagamihara, Kanagawa 252-0245, Japan
[*4]Department of Mechanical Engineering, Tokyo University of Science
6-3-1 Niijuku, Katsushika, Tokyo 125-8585, Japan
[†]Corresponding author

To achieve continuous frail care in the daily lives of the elderly, we propose AHOBO, a frail care robot for the elderly at home. Two types of support systems by AHOBO were implemented to support the elderly in both physical health and psychological aspects. For physical health frailty care, we focused on blood pressure and developed a support system for blood pressure measurement with AHOBO. For psychological frailty care, we implemented reminiscent coloring with the AHOBO as a recreational activity with the robot. The usability of the system was evaluated based on the assumption of continuous use in daily life. For the support system in blood pressure measurement, we performed a qualitative evaluation using a questionnaire for 16 subjects, including elderly people under blood pressure measurement by the system. The results confirmed that the proposed robot does not affect the blood pressure readings and is acceptable in terms of ease of use based on subjective evaluation. For the reminiscent coloring interaction, subjective evaluation was conducted on two elderly people under the verbal fluency task, and it has been confirmed that the interaction can be used continuously in daily life. The widespread use of the proposed robot as an interface for AI that supports daily life will lead to a society in which AI robots support people from the cradle to the grave.

## 1. Introduction

The older we get, the more help we need in our daily lives. The world's population is aging rapidly. Japan is a super-aging society with the highest rate of aging globally, where robots, AI, and IoT are expected to be used to support the elderly for the rest of their lives.

In Japan, the incidence of unintentional home accidents among the elderly is increasing. To help prevent accidents at home, care for various functional changes and losses of capacity due to aging and frailties is needed. Frail care requires continuous support in daily life.

Frail is defined as a condition in which one or more of the three functions – physical, psychological, and social - are lost or affected [1]. Although these functional losses may affect each other, physical and psychological frailty care is necessary in home situations.

In this study, we focused on blood pressure as an indicator of physical frailty. More than 80% of deaths at home are among the elderly, and an analysis suggests that the majority of deaths are possibly due to blood pressure changes [2]. Therefore, awareness among elderly of their blood pressure status at all times is important for taking care of their health condition and preventing accidents at home.

In addition, we focus on recreational activities with a robot that incorporates personal information. As a recreation, we focused on training to maintain cognitive function. Maintenance of cognitive function is a common concern for many elderly people.

Thus, we propose a support robot, AHOBO, to realize continuous frailty care for elderly at home in their daily lives. We implemented two types of robot applications to support the physical and psychological frailty care of the elderly and verified the possibility of continuous use in daily life.

For physical frailty care, we propose a robot support system to help elderly people take care of their frailty by themselves. In the proposed system, the robot will instruct elderly people on appropriate blood pressure measurement habits and methods and advise them on improving their health based on the visualized measurement data.

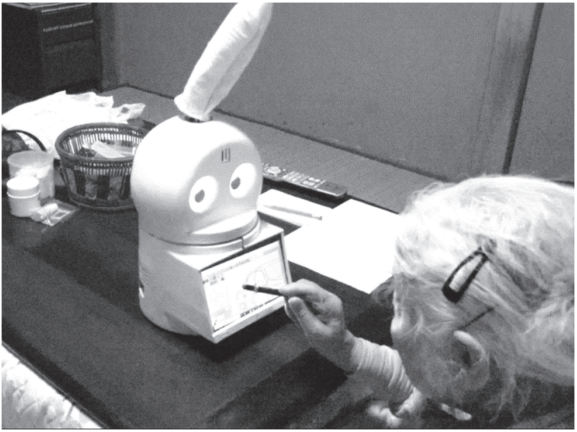

**Fig. 1.** Frailty care robot AHOBO and its experimental situation in the recreation with reminiscent coloring.

We examined its usability and also the influence of robotic blood pressure measurement support on blood pressure readings.

For psychological frailty care, we implemented reminiscent coloring with AHOBO as a recreational activity. The usability of the system was evaluated based on the assumption of continuous operation in daily life. A recreational activity scene between the AHOBO and an elderly person is shown in **Fig. 1**.

In Section 2, we describe the health habits and robots for frailty care. In Section 3, we describe the frailty care robot AHOBO and its applications developed in this study. As examples of applications and their evaluations, we describe blood pressure measurement support for physical frail care in Section 4, and a cognitive application using reminiscent coloring for psychological frailty care in Section 5.

## 2. Healthy Habits to Prevent Frailty and Robots for the Elderly

### 2.1. Frail Care and Robotic Applications for the Elderly

Frailty is a state of increased vulnerability to health maintenance due to various functional changes or loss of ability with aging, and is similar to ME-BYO (presymptomatic disease), which is well known in Japan. Frailty is reversible and can be restored to a healthy state with appropriate intervention. Therefore, it is important to establish a system for early detection of frailty.

Frail is a condition in which one or more of the three functions – physical, psychological, and social – are lost or affected [1]. With the rapid aging of the population, robots are expected to be used for voluntary frail care of the elderly.

For physical frailty care, both physical abilities and health are to be considered. Robotic rehabilitation support has been proposed for the care of physical abilities [3]. Health support through interaction with robots is effective [4], and a personal healthcare companion robot Mabu was released [5]. Mabu can monitor the health status of patients with lifestyle-related diseases and inform them about their medication intake timings through interaction.

For psychological frail care, daily conversation and information support by robots have been proposed, and their effectiveness for elderly people living alone has been confirmed [6]. As for the care of social frailty by robots, Telenoid [7] and OriHime [8] have been confirmed to connect people with each other.

Furthermore, since the three factors of frailty are closely related, robotic applications have also been confirmed to affect multiple factors. Paro, a seal-like mental commitment robot, has been confirmed to have both psychological and social effects [9, 10]. Regarding the combination of physical and social frailties, Kobayashi et al. [11] developed a robot that can perform stroke sign detection using SNS agency robots.

With regard to physical and psychological frailty, a health exercise support system using a partner robot has been proposed as an application that focuses on maintaining physical ability and its effects on physical activity and motivation have been confirmed [12].

In this study, we focused on maintaining health through lifestyle management and conversation during recreation for physical and psychological frailty. These factors are easier for the elderly to manage on their own.

### 2.2. Frailty Prevention and Blood Pressure in Elderly

Health conditions and blood pressure are prone to fluctuations, which can lead to accidents at home. According to the demographic survey of the Ministry of Health, Labour and Welfare (MHLW) in Japan, the number of unintentional fatal accidents in the elderly (aged 65 years and older) at home has been increasing both in terms of the number of accidents and the percentage of all ages [1]. Self-monitoring of blood pressure not only prevents frailty, but also helps prevent accidents [13].

Blood pressure measurement as part of daily health management is generally not customary and is not used to ascertain daily fluctuations. Therefore, there is a need for guidance in acquiring correct measurement habits and methods [14].

In addition, it has been pointed out that the visualization of health data such as blood pressure can change lifestyle habits. It has been shown that in an environment designed for internal corporate use, the visualization of health data can change behavior to improve health [15].

Based on the above, we focused on blood pressure as a lifestyle indicator to support the health aspect of physical frailty at home.

### 2.3. Recreation with Robots at Home

How to convey information is important for AI and robots that support our daily lives. The importance of casual verbal and non-verbal interactions with robots at home has been pointed out [16–18]. Robots linked to IoT

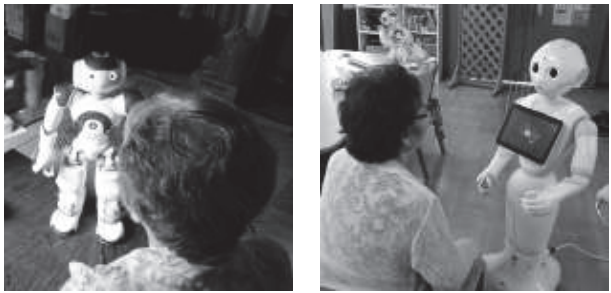

**Fig. 2.** Interview scene about home robot. The left photo is with NAO at the elderly people's home. The right is with Pepper in the elderly facility.

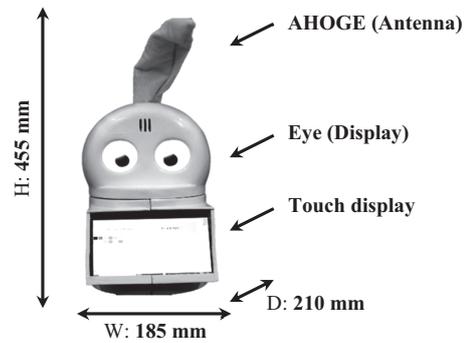

**Fig. 3.** Appearance of frailty care robot "AHOBO." Eye part and AHOGE (Antenna Hair-type Object for Generating Empathy) part give emotional expressions.

environments potentially will be used to observe elderly people at home [6].

As for the psychological effect in conversations with robots at home, it has been confirmed that the use of personal information leads to more smiles and friendliness [19].

Based on the above, we focus on robot applications that incorporate personal information in recreation and conversation to realize psychological frailty care. As a recreational activity, we focused on training to maintain cognitive function. The maintenance of cognitive function is a concern in many elderly people and training for this purpose is widely known as "NOTORE (brain training)" in Japan [20].

## 3. Elderly Support Robot AHOBO and its Applications for Physical and Psychological Frailty Care

We developed AHOBO as a robot to support individuals in their daily lives at home to prevent frailty in elderly people. Focusing on the health aspect of physical frailty, we built a robotic support system where AHOBO teaches appropriate blood pressure measurement habits and methods and advises on health improvement based on visualized measurement data [13]. In addition, focusing on psychological frailties, we have built a recreational application that incorporates personal information with conversations with AHOBO.

### 3.1. Interview Survey for Robots to Support Elderly Individual

We visited the elderly facility and elderly people's homes and interviewed elderly people in their 60s to 90s about home robots. The interview scene is shown in **Fig. 2**. A summary of the interviews is as follows.

- It is not desirable for elderly people to keep too large robots at home.

- Elderly people feel uncomfortable when the robot's hands are different from those of humans.

- Elderly people feel uncomfortable with bipedal robots because the robots are not stationary to maintain their posture.

Based on the above, we established the following requirements for robots to be placed in the homes of elderly people.

- Desktop size.

- It does not have a walking or moving mechanism. It can be placed in a fixed position.

- It has minimalistic emotional functions to be perceived as an intelligent character.

### 3.2. Frailty Support Robot AHOBO with Emotional Expression

We developed an AHOBO as a robot to support individuals in their daily lives at home to prevent frailty in elderly people. The appearance of AHOBO is illustrated in **Fig. 3**.

Based on the interview survey, the proposed robot is not a bipedal robot, but a stable and stationary robot with minimal moving parts and no fine movement.

Motif as an object is used to provide familiarity for the elderly people. Therefore, we decided to create a tabletop model with a bun motif. The proposed robot has a width of 185 mm, depth of 285 mm, and height of 455 mm.

The robot is capable of: interaction using a touch display, speaking, and expressing emotions through eyes and AHOGE for friendly communication.

### 3.3. Emotional Expression of AHOBO Using its Eyes and AHOGE

To provide friendly communication, AHOBO has an emotional expression function using its eyes and an antenna hair-type object for generating empathy (AHOGE). AHOGE is an emotional expression element that does not imitate the human style but uses the robot's own body [21]. The AHOGE part developed for AHOBO expresses four types of emotions: joy, anger, sadness, and surprise, with two degrees of freedom. **Fig. 4** shows the emotions expressed by the display of the eyes, and **Fig. 5** shows the emotions expressed by AHOGE. These emotional expressions with the eyes and with the AHOGE are combined in a coordinated/confrontational manner, depending on the situation.

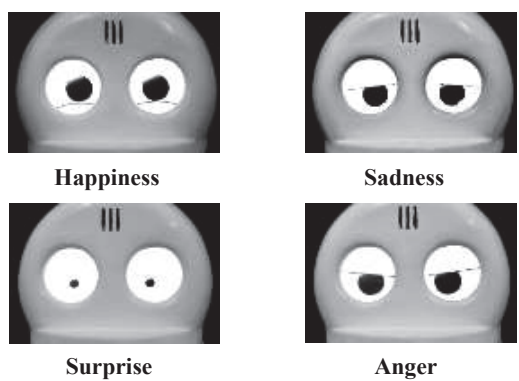

**Fig. 4.** Four emotional expression with eyes of AHOBO.

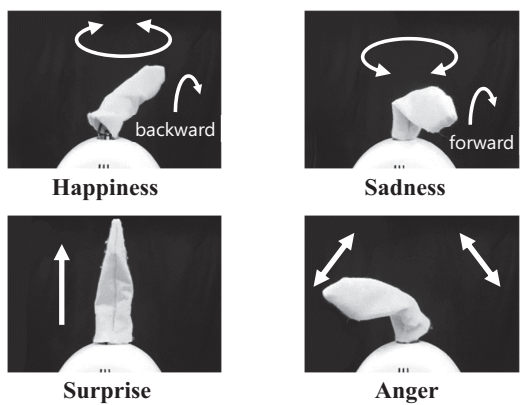

**Fig. 5.** Four emotional expression with AHOGE of AHOBO.

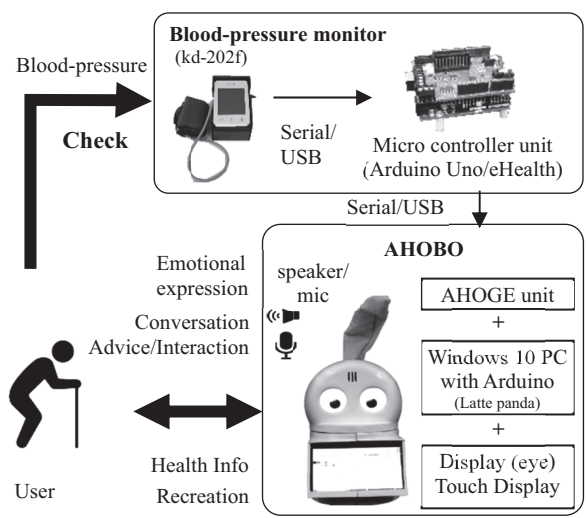

**Fig. 6.** Application for blood-pressure measurement with AHOBO.

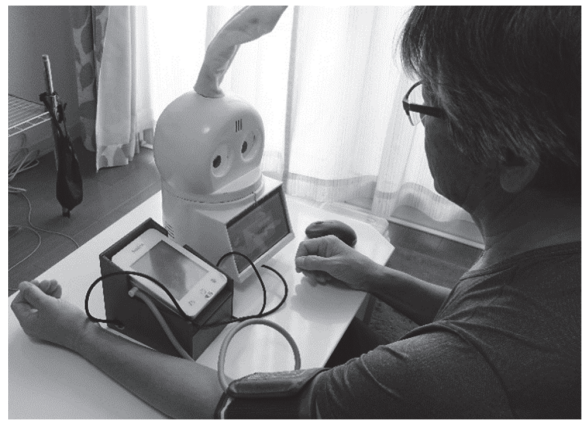

**Fig. 7.** Application for blood-pressure measurement with AHOBO.

### 3.4. Physical Frailty Support with Blood Pressure Measurement

Focusing on physical frailty, we proposed and constructed a robotic support system with a support robot where the support robot teaches appropriate blood pressure measurement habits and methods and provides advice for health improvement based on visualized measurement data.

The support robot encourages users to periodically measure their blood pressure in daily lives. AHOBO was prepared as a support robot and built into the system. An overview and the appearance of the proposed system using AHOBO are shown in **Figs. 6** and **7**. We used a KD-202F sphygmomanometer (Kodea Co., Ltd.) with an accuracy of $\pm 3$ mmHg.

During blood pressure measurement, the robot provides voice instructions on the proper use of a blood pressure monitor. After the measurement, the blood pressure information, visualized as a graph, is presented on the display, and the robot gives voice advice on how to improve the lifestyle according to the estimated health condition based on the measured blood pressure value. The robot creates emotional expressions during voice instructions and advice.

The classification of blood pressure values used in the proposed system was based on the Guidelines for the Treatment of Hypertension 2014 [22]. If the blood pressure is above pre-hypertension (defined as a systolic and/or diastolic blood pressure $\geq 140/90$), advice on eating and exercise habits are provided. Regarding dietary habits, the system recommends a low-sodium diet and a diet consisting mainly of vegetables to reduce high blood pressure. Regarding exercise, the system recommends a short indoor exercise program.

### 3.5. Psychological Frailty Support with Reminiscent Coloring

To support psychological frailty, a recreational application with conversation using AHOBO was proposed and implemented. The outline of the proposed system with AHOBO is shown in **Fig. 6**, and the actual usage scenario is shown in **Fig. 1**.

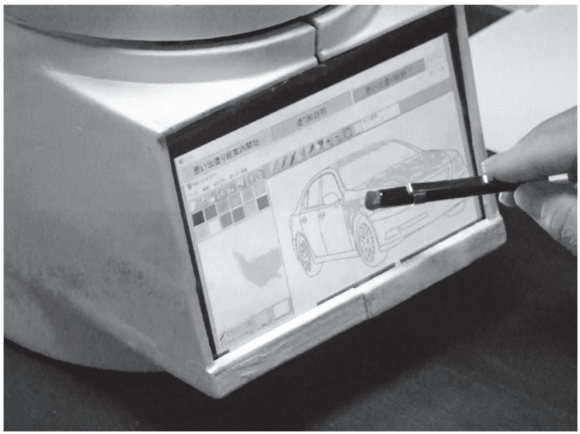

**Fig. 8.** Application for blood-pressure measurement with AHOBO.

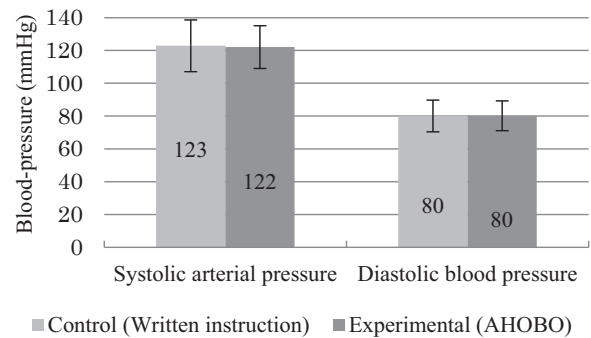

**Fig. 9.** Average of blood-pressure value in the experimental condition and the control condition.

In the proposed system, we prepared a training program to support the maintenance of cognitive functions as a recreational activity in conversation. We used reminiscent coloring as a training program, which is an intervention method that combines a coloring book and a reminiscence method often used for the prevention of nursing care for the elderly. By painting line drawings related to personal memories, memory activation and psychological stability can be expected [23].

In the proposed system, the user paints a line drawing on a touch display on the body of AHOBO, following voice instructions and conversations with AHOBO. As an application for coloring, PaintEmotion is used [24], which can express watercolor-like drawings on the display.

## 4. Verification of Effect of Blood Pressure Measurement Support Robot on Blood Pressure

For daily use of the proposed blood pressure measurement support robot, it is necessary that it should not affect the blood pressure measurements. To verify the influence of the proposed blood pressure support system on blood pressure readings, we compared blood pressure readings under two conditions: measurement by the proposed system and measurement by written instructions. In addition, we conducted a qualitative evaluation using a questionnaire.

### 4.1. Experimental Procedure

The experimental setup is illustrated in **Fig. 7**. The experimental procedure was as follows:

Step 1) Explain the experiment to the participant.

Step 2) Measure the blood pressure according to each condition.

In this verification, the following two conditions were set up: In the experimental condition "robot instruction (proposed)" the participant measures the blood pressure according to the voice instruction by the robot, and the measurement results are presented on the display of the robot or on the display beside the robot. In the control condition "written instructions," the participant measures the blood pressure according to the written instructions, and the results are presented on the display. The order of the experiments for each condition was randomly determined.

Step 3) Ask the participants to answer the questionnaire.

The questionnaire contains three subjective evaluations: "Are you nervous with the instructions?," "Are the instructions easy to understand?," and "Is the system easy to use?" The participants rated it on a 5-point scale.

Step 4) Repeat steps 2 and 3 with the other conditions.

During the experiment, AHOBO expresses happiness with eyes and the AHOGE while giving each instruction, and expresses happiness or sadness in accordance with the blood pressure value.

This experiment was conducted on a total of 16 subjects: five in their 20s, four in their 50s, five in their 60s, and two in their 70s.

In addition, the following data were used as a baseline: blood pressure measurement according to written instructions without AHOBO performed on both occasions (Baseline). There were 16 subjects: 13 subjects in their 20s, one in their 40s, one in their 60s, and one in their 70s.

### 4.2. Results of Blood Pressure Measurement

The average blood pressure measurement results in the experimental and control conditions are shown in **Fig. 9**. The average blood pressure measurement results under the control conditions (baseline) are shown in **Fig. 10**. There was no significant difference in the mean values of the two blood pressure measurements, either in the results of comparing the experimental condition (robot instructions) with the control condition (written instructions) (**Fig. 9**) or in the results of repeating the control condition (written instructions) twice (**Fig. 10**).

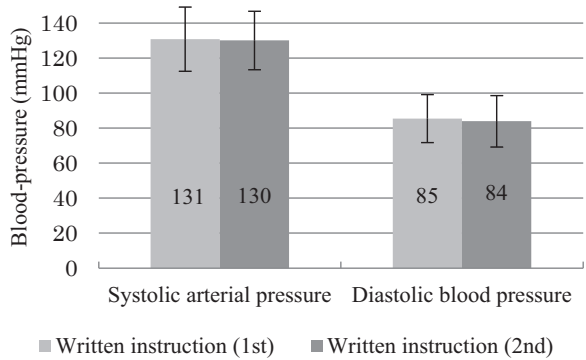

**Fig. 10.** Average of blood-pressure value in the control conditions (**Baseline**).

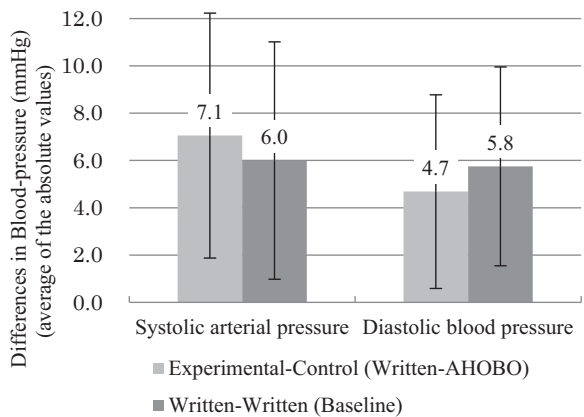

**Fig. 11.** Differences in individual blood pressure values. This is the average of the absolute values of the differences in individual blood pressure values over the two measurements.

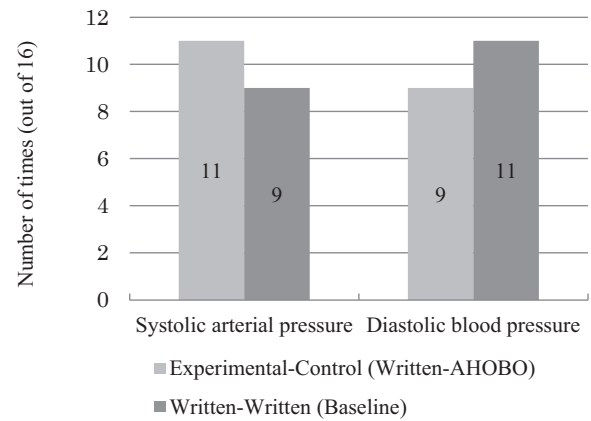

**Fig. 12.** Number of times (out of 16) that the difference between the two measurements exceeded the measurement error of the sphygmomanometer ($\pm 3$ mmHg).

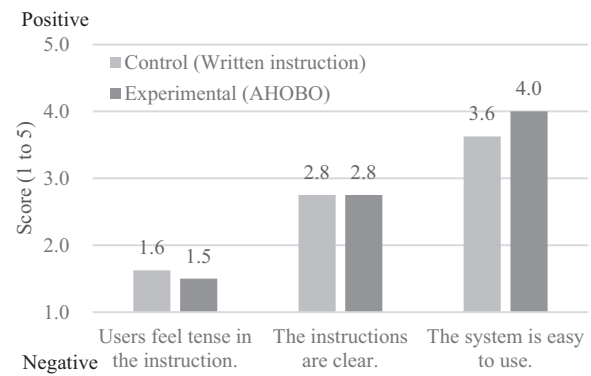

**Fig. 13.** Average of the evaluated value in the questionnaire.

Focusing on the differences in individual measurements, the average of the absolute values of the differences in individual blood pressure values over the two measurements are shown in **Fig. 11**. The number of times (out of 16) that the difference between the two measurements exceeded the measurement error of the sphygmomanometer ($\pm 3$ mmHg) is shown in **Fig. 12**. There was no significant difference between the results of comparing the experimental condition (robot instructions) with the control condition (written instructions) and the results of repeating the control condition (written instructions) twice, both in the mean absolute value of the error and in the number of times the measurement error precision was exceeded. In addition, in **Fig. 11**, it can be confirmed that the blood pressure values are not affected by the robot as the difference in the mean of the absolute values of the differences in the individual blood pressure measurements for both systolic and diastolic blood pressure was within the measurement error precision ($\pm 3$ mmHg).

### 4.3. Results of Subjective Questionnaire

The average of the questionnaire results is shown in **Fig. 13**. Although the experimental condition (AHOBO instruction) was more positive than the control condition (Written instruction) in the items "Users feel tense in the instruction." and "The system is easy to use.," there has been no significant difference.

In the experimental scene, conversations were often repeated due to misinterpretation of speech recognition. Improving speech recognition may improve the results of the subjective questionnaires.

## 5. Evaluation of Interaction with Reminiscent Coloring

A recreational application with reminiscent coloring using AHOBO was proposed and implemented to support psychological frailty. In order to confirm the effectiveness of the proposed system, it is necessary to use it for a long period with many people. As a preliminary step, we aim to investigate the possibility of the continuous use of AHOBO by elderly people in this study. We conducted a verbal fluency task in reminiscent coloring interaction with AHOBO at home for two elderly people and evaluated its usability in a short period of time.

## 5.1. Verification of Ease of Use of Reminiscent Coloring with AHOBO by Elderly People

To verify the usability of the proposed AHOBO and its application with reminiscent coloring, we conducted an interaction experiment with elderly people in their own homes and evaluated the usability and its effect on word fluency.

Subjective questionnaires were used to evaluate the usability of the system. The elderly were asked to rate on a 5-point scale "the ease of use of the proposed system," "the ease of coloring," and "the ease of understanding the explanations." Verbal fluency is the ability to calculate words related to a category. To evaluate this, we used "animal" as a specific word indicating a category in this study, referring to Tanaka et al. [23].

Interactions with AHOBO were conducted for one day. The verbal fluency tasks were conducted at four different times: (i) three days before the interaction experiment, (ii) just before the experiment on the day of the interaction experiment, (iii) just after the interaction experiment, and (iv) three days after the interaction experiment.

The procedure of the interaction experiment on the day is as follows:

Step 1) Conduct verbal fluency assessment ((ii) just before the experiment).

Step 2) Participants do a reminiscent coloring while interacting with AHOBO.

Step 3) Evaluate language fluency ((iii) immediately after).

Step 4) Conduct a subjective questionnaire on the ease of use.

We conducted the above tests on two elderly people over 65 years of age for a total of one week at their homes. During the experiments, we interviewed them about their living conditions.

## 5.2. Experimental Results

The results of the verbal fluency task are shown in **Fig. 14**, and the results of the subjective questionnaire are shown in **Fig. 15**. There were no significant changes in the verbal fluency task.

As a result of the interviews during the one-week experiment, no significant changes were observed in the living conditions of the two elderly people.

## 5.3. Discussions

The results of the subjective questionnaire shown in **Fig. 14** indicate that the two elderly people scored above 3, which is the average value of the evaluation, for all items. From this result, we conclude that the elderly can use AHOBO in their living environment.

In addition, the results of the interviews during the one-week experiment showed no significant changes in the living conditions of the two elderly people, and we concluded that the system could be used continuously without any problems.

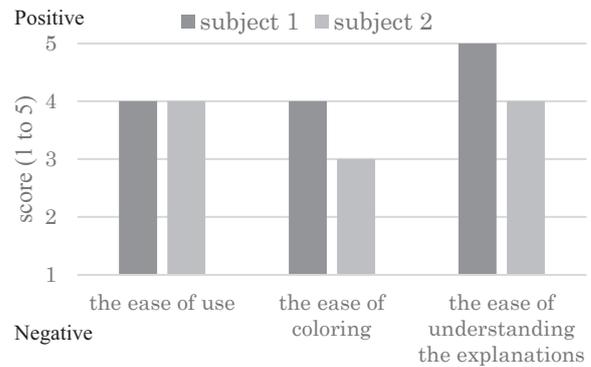

**Fig. 14.** Score of verbal fluency task (for two subjects over 65 years old).

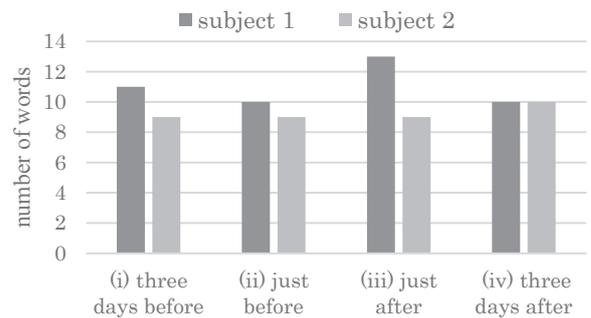

**Fig. 15.** Score of subjective questionnaires (for two subjects over 65 years old).

## 6. Conclusion

In this study, we developed a support robot AHOBO to realize continuous frailty care in daily life for the elderly at home. We also proposed robot support systems to support the elderly in terms of both psychological and physical health aspects. For physical frailty care in health, we proposed a blood pressure measurement support. For psychological frailty care, we implemented reminiscent coloring with AHOBO as a recreational activity that incorporates personal information in interactions with the robot. The usability of the system in daily life was verified based on the assumption of continuous use in daily life. As a result, we confirmed that the robot does not affect the blood pressure readings and is acceptable in terms of ease of use. Based on these results, we conclude that the proposed robot and its application can realize continuous frailty care in daily life for the elderly at home.

The proposed robot is expected to contribute to health maintenance, cognition, etc., through long-term daily use, and its evaluation will be our future work. In the next phase of the study, we plan to examine the effects of AHOBO on cognitive function by preventing physical and social frailty.

The widespread use of the proposed robot, as an interface for AI, that supports elderly in their daily lives and communicates with them, will lead to a society where AI robots will support people from the cradle to the grave.

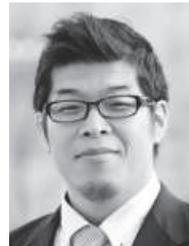

**Name:**
Yoichi Yamazaki

**Affiliation:**
Department of Home Electronics, Faculty of Creative Engineering, Kanagawa Institute of Technology

**Address:**
1030 Shomo-ogino, Atsugi, Kanagawa 243-0292, Japan
**Brief Biographical History:**
2009-2013 Assistant Professor, Kanto Gakuin University
2012-2013 Visiting Researcher, Beijing Institute of Technology
2013- Associate Professor, Kanagawa Institute of Technology
**Main Works:**
● Y. Yamazaki, Y. Hatakeyama, F. Dong, K. Nomoto, and K. Hirota, "Fuzzy Inference based Mentality Expression for Eye Robot in Affinity Pleasure-Arousal Space," J. Adv. Comput. Intell. Intell. Inform., Vol.12, No.3, pp. 304-313, 2008.
● K. Takeuchi, Y. Yamazaki, and K. Yoshifuji, "Avatar Work: Telework for Disabled People Unable to Go Outside by Using Avatar Robots," Companion of the 2020 ACM/IEEE Int. Conf. on Human–Robot Interaction (HRI '20), pp. 53-60, 2020.
**Membership in Academic Societies:**
● The Robotics Society of Japan (RSJ)
● Japan Society for Fuzzy Theory and Intelligent Informatics (SOFT)

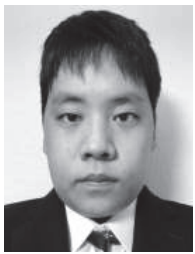

**Name:**
Masayuki Ishii

**Affiliation:**
OKUBO GEAR Co., Ltd.

**Address:**
3030 Kamiechi, Atsugi, Kanagawa 243-0801, Japan
**Brief Biographical History:**
2018 Received the B.Eng. from Department of Home Electronics, Kanagawa Institute of Technology
2018- Joined OKUBO GEAR Co., Ltd.

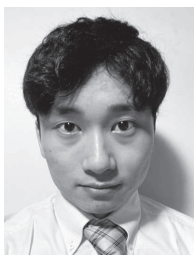

**Name:**
Takahiro Ito

**Affiliation:**
CMK Products Corporation

**Address:**
1-1-11 Tanashioda, Chuo, Sagamihara, Kanagawa 252-0245, Japan
**Brief Biographical History:**
2018 Received the B.Eng. from Department of Home Electronics, Kanagawa Institute of Technology
2018- Joined CMK Products Corporation

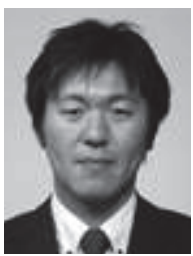

**Name:**
Takuya Hashimoto

**Affiliation:**
Department of Mechanical Engineering, Tokyo University of Science

**Address:**
6-3-1 Niijuku, Katsushika, Tokyo 125-8585, Japan
**Brief Biographical History:**
2009-2012 Assistant Professor, Tokyo University of Science
2013-2015 Assistant Professor, The University of Electro-Communications
2016- Junior Associate Professor, Tokyo University of Science
**Main Works:**
● T. Hashimoto, S. Hitramatsu, T. Tsuji, and H. Kobayashi, "Development of the Face Robot SAYA for Rich Facial Expressions," 2006 SICE-ICASE Int. Joint Conf., pp. 5423-5428, doi: 10.1109/SICE.2006.315537, 2006.
**Membership in Academic Societies:**
● The Institute of Electrical and Electronics Engineers (IEEE)
● The Japan Society of Mechanical Engineers (JSME)
● The Robotics Society of Japan (RSJ)
● Human Interface Society